\pdfoutput=1
\documentclass[11pt,a4paper]{article}
\usepackage{times,latexsym}
\usepackage{url}
\usepackage[T1]{fontenc}

\usepackage{enumitem}

\usepackage{pgfplots}
\usepgfplotslibrary{groupplots}
\usepackage{listings}

\usepackage{xcolor}
\usepackage[acceptedWithA]{tacl2021v1}

\usepackage{xspace,mfirstuc,tabulary}

\newcommand{\rev}[1]{\textcolor{black}{#1}}

\newif\iftaclinstructions
\taclinstructionsfalse 
\iftaclinstructions
\renewcommand{\confidential}{}
\renewcommand{\anonsubtext}{(No author info supplied here, for consistency with
TACL-submission anonymization requirements)}
\newcommand{\instr}
\fi

\iftaclpubformat 

\else

\fi

\usepackage{comment}

\usepackage{graphicx}
\usepackage{scalerel} 

\usepackage{adjustbox}

\usepackage{bm} 
\usepackage{mathtools} 
\usepackage{amsfonts} 
\usepackage{nicefrac} 

\usepackage{enumitem} 

\usepackage{multirow}
\usepackage{booktabs}
\usepackage{colortbl}
\usepackage{arydshln} 
\usepackage{tabularx}
\usepackage{threeparttable} 

\usepackage{pgfplots}
\usetikzlibrary{pgfplots.fillbetween,patterns}
\usepackage{pgfplotstable}
\usepgfplotslibrary{groupplots}
\usepackage{scalerel} 
\usepackage{stfloats} 
\usepackage{caption}
\usepackage{subcaption}

\usepackage{tikz}
\usetikzlibrary{arrows,automata,calc,shapes, positioning,shadows,shadows.blur,shapes.geometric}

\usepackage{listings}
\usepackage[most]{tcolorbox}

\usepackage{pifont}

\usepackage[dash,dot]{dashundergaps}

\usepackage{cleveref}
\crefformat{section}{\S#2#1#3} 
\crefformat{subsection}{\S#2#1#3}
\crefformat{subsubsection}{\S#2#1#3}
\crefformat{appendix}{\S#2#1#3}

\usepackage{fontawesome5}

\usepackage{tcolorbox}
\tcbuselibrary{breakable,xparse,skins}

\definecolor{orders-color}{HTML}{D79B00}
\definecolor{judgments-color}{HTML}{6C8EBF}
\definecolor{maxims-color}{HTML}{917FB3}
\definecolor{maxims-titles-color}{HTML}{C9A7EB}
\definecolor{chart-background}{HTML}{F0F0F0}
\definecolor{avg-color}{HTML}{E5ECF6}
\definecolor{ita-green-color}{HTML}{4CAF50}
\definecolor{ita-white-color}{HTML}{FDFDF4}
\definecolor{ita-red-color}{HTML}{F44034}
\definecolor{blue}{HTML}{7A7EFF}
\definecolor{bg}{gray}{0.95}

\title{{\sc Finch}: \rev{Prompt-guided Key-Value Cache Compression}\\for 
Large Language Models}

\author{
Giulio Corallo
  \\
  SAP Labs, France
  \\
  EURECOM, France
  \\
  \texttt{giulio.corallo@sap.com}
  \And
Paolo Papotti
  \\
EURECOM, France
  \\
  \texttt{papotti@eurecom.fr}
}

\date{}

\begin{document}

\maketitle

\begin{abstract}
Recent large language model applications, such as Retrieval-Augmented Generation and chatbots, have led to an increased need to process longer \rev{input} contexts.
However, this requirement is hampered by inherent limitations. Architecturally, models are constrained by a context window defined during training. Additionally, processing extensive texts requires substantial GPU memory. We propose a novel approach, \rev{{\sc Finch}}, \rev{to compress the input context by} leveraging the pre-trained model weights of the self-attention. Given a prompt and a long text, \rev{{\sc Finch}} iteratively identifies the most relevant Key (K) and Value (V) pairs over chunks of the text conditioned on the prompt. Only such pairs are stored in the KV cache, which, within the space constrained by the context window, ultimately contains a compressed version of the long text.
Our proposal enables models to consume large inputs even with high compression (up to \rev{93x}) while preserving semantic integrity without the need for fine-tuning. 
\end{abstract}

\section{Introduction}
Large Language Models (LLMs), built upon the Transformer architecture, have delivered breakthroughs in numerous applications. 
With their generalization and reasoning capabilities, models such as ChatGPT have revolutionized fields where extensive \rev{input} prompts 
are necessary for generating precise responses, such as Retrieval-Augmented Generation, Chain-of-Thought, conversational chatbots, and In-Context Learning~\citep{rag20,wei2022chain,dong2022survey}. 

However, the use of LLMs in production is limited by their increasing requests in terms of GPU memory~\citep{dettmers2024qlora}.
First, as the computational complexity grows along with the size of the models, their memory consumption increases. 
Second, this issue becomes more pronounced when LLMs process larger inputs, as demanded by their ever-increasing context size. 
Third, the Key-Value (KV) cache mechanism, typically employed by LLMs to speed up the generation process, prioritizes efficiency by retaining and reusing previously computed KV vectors during attention computation, bypassing 
re-calculations at each token
generation step~\citep{kaiser2017learning}. Nevertheless, this solution comes with the trade-off of increased memory consumption.\footnote{\rev{It has been reported that OPT-175B (with batch size 128 and sequence length 2048) consumes 325 GB of memory, but its KV cache requires 950 GB~\citep{liu2023scissorhands}}.} 

To offer more efficient solutions to operate these models, it has been proposed to \textit{compress} \rev{input} prompts, exploiting the redundancy in natural language~\citep{goyal2020power}. By preserving critical token information while compressing less crucial details, these models reduce the context in a compact description, without noticeably 
degrading the functional accuracy~\citep{mu2024learning}. Compression also enables the LLMs to process large \rev{inputs} that do not fit the model's context size. However, most of these models require a training/fine-tuning process or a large number of calls to an external model for the compression~\citep{JiangWLYQ23}.

We revisit the LLMs' generative inference mechanism to deal with the memory constraint problem and the limitations of current solutions \rev{in processing large inputs}. We propose a novel approach targeting the reduction of the KV cache memory footprint while avoiding resource-intensive retraining or fine-tuning processes. Drawing insights from the patterns inherent in attention modules, and guided by the understanding that not all attention modules engage with every token, our solution compresses the cached vectors, leading to a reduction in memory usage and efficient text generation.

\begin{figure*}[!t]
    \centering
   \begin{adjustbox}{width=.95\linewidth}
\includegraphics{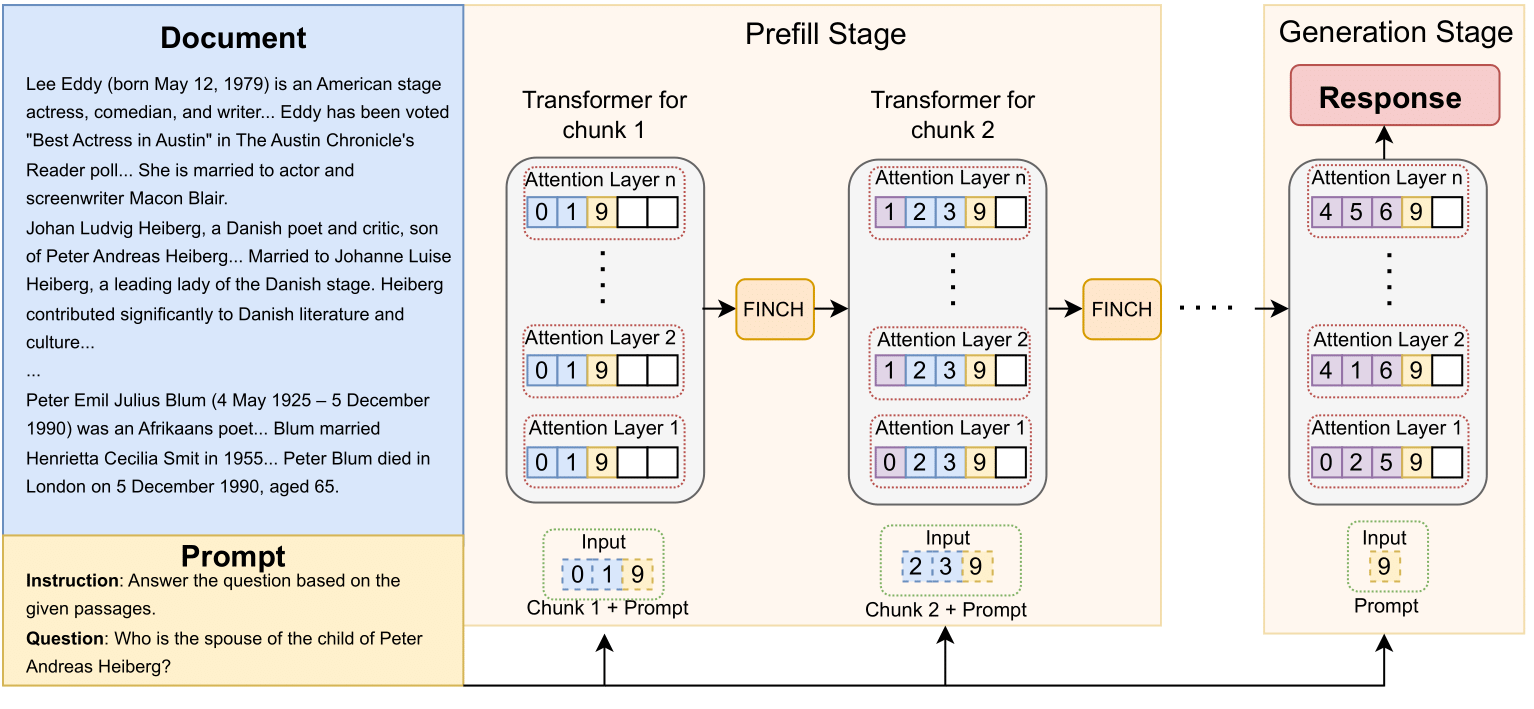}
   \end{adjustbox}
    \caption{Overview of \textsc{Finch}. An input document is larger than the model context and thus is processed in chunks. \rev{At each step in the Prefill stage, \textsc{Finch} 
    sequentially consumes a document chunk (two dashed border squares, blue background), alongside the input prompt (one dashed border square, yellow background) as depicted at the bottom. At each step, it 
     processes the 
    key, value pairs (solid squares in the transformer) and identifies the most relevant to the prompt. It then carries them to the cache processing the next chunk (where they appear with a violet background). In the Generation stage, the model synthesizes a response that is informed by the compressed cached information from the entire document. The white square is the space reserved for producing output tokens in the Generation stage.}}
    \label{fig:archi}
\end{figure*}

Our approach, termed \textsc{Finch},\footnote{\textit{Finch} is a small and quick bird, known for its chirp – a complex language for a small animal.} facilitates faster generative inference through adaptive KV cache compression \rev{in the Prefill stage}. Figure~\ref{fig:archi} shows how a long document and the input prompt are processed with a model context size that cannot fit the entire input.  At every step, a document chunk is processed. \textsc{Finch} uses the attention information \rev{between the prompt and the document chunk} to identify the 
\rev{most relevant}
KV pairs across different layers.
This information then is stored in the KV cache for the processing of the next \rev{input} chunk. Our approach dynamically selects what to keep in the KV cache's memory, effectively keeping its footprint constrained, until the Generation stage produces the response. 

\textsc{Finch}, 
incrementally feeds the KV cache with the compressed context without any learning or external summarization module; it can be used in a plug-and-play fashion with any decoder-based model. The 
compression rate is specified by setting the \rev{target} size of the KV cache as an input parameter constrained by the model context size.




Even with high compression ratios, our method ensures that the 
correctness of the model response 
is preserved. 
We test \textsc{Finch} on two popular benchmarks covering tasks in question answering, summarization, code \rev{completion}, \rev{synthetic tasks} and few-shot learning. Compared against the original LLM (without compression) over the SQuAD v2 benchmark~\citep{rajpurkar-etal-2018-know}, \textsc{Finch} achieves comparable generation quality at \rev{2.35x} compression and 90\% of the reference accuracy score at \rev{3.76x} compression, while being faster in terms of end-to-end execution times in most cases.
When compared to the state-of-the-art compression method \rev{LongLLMLingua~\citep{jiang2024longllmlingua}} 
\textsc{Finch} \rev{reports the best quality scores in most of the tasks in LongBench~\citep{bai2023longbench}, both with Llama 2 and Mistral~\citep{touvron2023llama2, jiang2023mistral}}. 
\rev{Our method achieves a compression range of 2x to 93x across various tasks, consistently outperforming a truncation baseline in most experiments. Remarkably, \textsc{Finch} even surpasses the performance of the LLMs operating with the full, uncompressed context in certain cases. Finally, in question answering tasks, we also include a RAG baseline, and our method outperforms it in 10 out of 12 experiments.}

\section{Related Work}
We position our work w.r.t. two main topics.
First, we discuss strategies for improving computational efficiency, i.e., making LLMs accessible for real-time applications or use on devices with limited resources. 
Second, we focus on attention patterns in LLMs, as our work shows that those contribute significantly towards optimizing the models to handle larger inputs in a limited context size.

\noindent \textbf{Efficiency Improvements in LLMs.} Methods targeting the reduction of inference and fine-tuning costs include models' modification, such as quantization~\citep{frantar2023optq,dettmers2022gptint} and model compression~\citep{FrantarA23}.
Other efforts enhance model efficiency for LLMs by eliminating redundant input words based on attention scores~\citep{goyal2020power} and compressing the input sequence by augmenting the encoding modules with pooling layers~\citep{Dai_NEURIPS20}. Proposed solutions also involve learning to skip layers in the transformer architecture~\citep{guan-etal-2022,zhou_NEURIPS20} or to select the most critical tokens for performance~\citep{huang-etal-2022}.
Other approaches pursue prompt compression, either by limiting the number of tokens that are processed in inference by learning special ``compressed'' tokens~\citep{mu2024learning,wingate-etal-2022-prompt,ge2024incontext} or by pruning and merging tokens~\citep{goyal2020power,modarressi-etal-2022}, e.g., learning thresholds for pruning unimportant ones~\citep{kim2022learned}. However, some of these strategies require an additional re-training or fine-tuning phase and others have been designed for encoder models and are not well suited for auto-regressive LLMs such as ChatGPT and Llama~\citep{touvron2023llama1, touvron2023llama2}. 
In contrast with such solutions, our approach condenses auto-regressive LLMs \rev{input} contexts during the Prefill stage by using the caching mechanism without model re-training and even faster inference. \rev{Finally, recent methods focus on optimizing the generation stage to improve efficiency~\citep{zhang2023ho,xiao2024efficient,han2024lminfinite,oren2024transformers,ren2024efficacy}. We leave to future work the study of how to use our prompt-guided token selection strategy in such approaches.}

\noindent \textbf{The Role of Attention.} Our work relies on self-attention to make the most relevant information in a context available in a concise manner.
The development of transformer models provoked studies to unravel the underlying mechanisms of self-attention, e.g., heads prominently pay attention to separator and adjacent tokens~\citep{clark-etal-2019-bert}. 
Our solution capitalizes on the attention mechanism structure to heighten inference efficiency by exploring the KV cache for the most important key, value pairs w.r.t. the given prompt.
Related work evaluates the informativeness of lexical units using a language model and drops less informative content for  compression~\citep{li2023unlocking,JiangWLYQ23,jiang2024longllmlingua}, for example by regarding tokens with lower perplexity as more influential in the inference process.
These techniques 
view LLMs as a compressor for world knowledge and work by further compressing information within prompts~\citep{deletang2024language}.
In contrast with these solutions, 
our approach instead optimizes the management of the KV cache during the Prefill stage without requiring a separate LLM.
\rev{Other approaches look at how to select the most important tokens in the Prefill stage, but, differently from our method that dynamically identifies the most important tokens, they rely on manually defined policies for token selection~\citep{ge2024model}.}

Finally, we focus on a plug-and-play solution for existing models, with an emphasis on limited computing resources. This is in contrast with other solutions that demand more devices to handle a very large input context
~\citep{liu2023ring}.

\section{Background}
Self-attention is foundational in transformer models~\citep{vaswani2017attention}, enabling language understanding and generation capabilities. 
Transformers learn the contextual relationships between words or subwords within a sentence.
Central to this mechanism are three types of vectors --- Queries ($\mathbf{Q}$), Keys ($\mathbf{K}$), and Values ($\mathbf{V}$) --- that are learned from the input embeddings. 
 \setlist{nolistsep}
    \begin{itemize}[leftmargin=*]
    \item \textbf{Queries ($\mathbf{Q}$)}: Represent the current word or token being processed, acting as a point of focus.
    \item \textbf{Keys ($\mathbf{K}$)}: Serve as identifiers, highlighting tokens in the sequence relevant to the query.
    \item \textbf{Values ($\mathbf{V}$)}: Correspond to the actual specific information carried by each token.
\end{itemize}
$$\text{Attention}(\mathbf{Q}, \mathbf{K}, \mathbf{V}) = \text{softmax}\left(\frac{\mathbf{Q}\mathbf{K^T}}{\sqrt{d_k}}\right)\mathbf{V}
$$
The self-attention mechanism computes the dot product of the \textbf{Query} with all \textbf{Keys} to determine their similarity. 
A softmax function normalizes these scores, creating a distribution that determines how much attention to allocate to each token. The output is a weighted sum of the \textbf{Values}.

In several NLP tasks, 
transformers generate a response sequence from a given context/document and a user prompt. Consider a sequence of tokens representing the context \rev{ \(\mathbf{x}^{\text{cont}} \in \mathbb{R}^{n^{\text{cont}}}\)} and a sequence of tokens representing the user prompt \rev{\(\mathbf{x}^{\text{que}} \in \mathbb{R}^{n^{\text{que}}}\)},
which may also include instructions,
the goal is to enable the model to generate a response sequence \rev{\(\mathbf{y} \in \mathbb{R}^{a}\)}. This process can be divided into two stages. 

\noindent \textbf{Prefill Stage.}
\rev{As a first step, both the context and the prompt sequence are concatenated to form the input sequence \(\mathbf{x} \in \mathbb{R}^{n}\), where:}
\rev{
\[
\mathbf{x} = \begin{bmatrix}
\mathbf{x}^{\text{cont}} \\
\mathbf{x}^{\text{que}}
\end{bmatrix}, \quad \text{and} \quad n = n^{\text{cont}} + n^{\text{que}}
\]
}
\rev{
This sequence is then embedded into an embedding matrix \(\mathbf{X} \in \mathbb{R}^{n \times d}\), where \(d\) denotes the embedding dimension and processed through multiple layers of multi-head self-attention and feed-forward networks, which operate in parallel across the sequence length and attention heads.} Each attention layer calculates and stores the corresponding Key and Value
$
\mathbf{K} \in \mathbb{R}^{n \times d}, \quad \mathbf{V} \in \mathbb{R}^{n \times d}
$
matrices in a cache for the sake of performance for the subsequent Generation stage.
In the transformer architectures, the $\mathbf{K}$ and $\mathbf{V}$ matrices encapsulate historical token information. Unlike other components of the transformer (e.g., feedforward layer or layer norm), which process current inputs independently of their past tokens, the $\mathbf{K}$ and $\mathbf{V}$ matrices are the only matrices 
that retain information from previously encountered tokens. Caching the Key and Value matrices for every layer for the context eliminates the necessity to recompute them for each new token generated.

\noindent \textbf{Generation Stage.}
In this step, 
the model iteratively generates new tokens. For each new token, the $\mathbf{q^{\text{new}}}, \mathbf{k^{\text{new}}}, \mathbf{v^{\text{new}}} \in \mathbb{R}^{d}$ are produced at each attention layer with the $\mathbf{k^{\text{new}}}$ and $\mathbf{v^{\text{new}}}$ vectors appended to the existing cache keys and values:
\rev{
\[
\mathbf{K} \leftarrow \begin{bmatrix}
\mathbf{K} \\
\mathbf{k^{\text{new}}} \\
\end{bmatrix}, \quad \mathbf{V} \leftarrow \begin{bmatrix}
\mathbf{V} \\
\mathbf{v^{\text{new}}} \\
\end{bmatrix}
\]
}
Self-attention uses this updated cache to compute attention. Thanks to the stored $\mathbf{K}$ and $\mathbf{V}$ matrices, the computational complexity is just $O(nd)$ as opposed to the approach without cache, which has a computational complexity of $O(n^2d)$. Finally, logits are generated and used to predict the next token in the vocabulary, 
e.g., with greedy decoding~\citep{vijayakumar2016diverse, shao2017generating}.

\section{Problem Formulation}
As discussed, $K$ and $V$ are the only matrices that retain information about previous tokens. 
We can therefore formulate the problem of compression as reducing the size of these two matrices during the {Prefill stage} and before the actual answer generation takes place. Specifically, we have to find $\mathbf{\Tilde{K}}$ and $\mathbf{\Tilde{V}}$ where $\mathbf{\Tilde{K}},\mathbf{\Tilde{V}} \in \mathbb{R}^{k \times d}$ such that two properties are satisfied:
\begin{itemize}[leftmargin=*]
    \item \textbf{Compression}: 
    the \rev{\textit{target tokens size}}
    $k$ 
    of the compressed $\mathbf{\Tilde{K}}, \mathbf{\Tilde{V}}$ matrices should be smaller than the sequence length $n^{\text{cont}}$ of $\mathbf{K^{\text{cont}}}, \mathbf{V^{\text{cont}}} \in \mathbb{R}^{n^{\text{cont}} \times d}$. 
    \item \textbf{Information retention}:
    the output $\mathbf{y} \in \mathbb{R}^{a}$ using $\mathbf{K}$,$\mathbf{V}$ matrices is similar to the output $\mathbf{\tilde{y}} \in \mathbb{R}^{a}$ obtained using $\mathbf{\Tilde{K}},\mathbf{\Tilde{V}}$, 
    expressed as:
    \begin{equation} \label{eq:similar_predictions}
    \displaystyle \min_{\mathbf{\tilde{K}}, \mathbf{\tilde{V}}} f(\mathbf{\tilde{y}}, \mathbf{y})
    \end{equation}
    where $f$ is a \rev{distance} function and its choice depends on the task at hand. For example, in question answering, the \rev{difference between} F1 scores \rev{for $\mathbf{\tilde{y}}$ and $\mathbf{y}$} might be used.
\end{itemize}

We also define 
the compression ratio $\sigma$ as: 
    \begin{equation*} \label{eq:compression}
        \sigma = \frac{n^{\text{cont}}}{k}
    \end{equation*}
In this work, we compress the context  $\mathbf{K^{\text{cont}}}, \mathbf{V^{\text{cont}}}$ matrices, \rev{according to the target tokens size $k$}, while conditioning on the user prompt. This decision is driven by the recognition that the integrity of the user prompt -- particularly its instructions for an instruction-tuned model -- plays a significant role in the answer generation~\citep{ouyang2022training}. 
Furthermore, in the 
tasks that we address in this work, the prompt is typically much shorter than the context, making its compression of limited value.

\section{Method}
Our approach aims at compressing contexts into a manageable form for LLMs, particularly when faced with extensive documents and the need to maintain computational efficiency. Our methodology is motivated by the following observation: the softmax of self-attention distributes attention across all elements to varying degrees,
effectively capturing a spectrum of contextual relationships in the data. We hypothesize that the "smooth" distribution of attention may include superfluous information for the given prompt at hand.

\subsection{Adaptive Key-Value Cache Compression}
As depicted in Figure~\ref{fig:archi}, \textsc{Finch} iteratively processes a document segmented into chunks, each evaluated in conjunction with a user prompt, and uses the self-attention to identify which  K,V pairs to keep in the cache. \rev{In analogy to the long-term memory involving the capacity to recall words, concepts, or numbers~\citep{chauvet2024memory}, we say that these pairs can act as the \textit{semantic memory} for the model.}
The document is reduced to its significant elements and processed in the Generation stage.

\vspace{0.5ex}
\noindent\textbf{Document Segmentation.}  
The transformer input is constrained by a context window defined during training, denoted as \rev{$n_{max}$}. 
Given the user specified {target tokens} size $k$ for the KV cache, \rev{{\sc Finch}} processes chunks using at most \rev{$m_{\text{max}}= n_{max} - k$ tokens}.\footnote{We ignore the user prompt size in this discussion as we assume it to be much smaller than \rev{the input document size}.} 
The input document is partitioned into chunks of size $m$, which value is constrained by $m_{\text{max}}$.  
At every Prefill step $i$, for $i>1$,
%
%
the K,V pairs from the previous step $i-1$ (the compressed chunk) are added into the tokens reserved for the $k$ target tokens.

This process introduces a trade-off between granularity and throughput. Smaller chunks enable finer granularity in processing, which is beneficial for certain tasks as we highlight in Section~\ref{sec:results}. Conversely, larger chunks (up to $m_{\text{max}}$) enhance throughput by reducing the number of sequential operations required, thus speeding up the Prefill stage. This trade-off 
is crucial for optimizing performance and is examined in our ablation study. 

\begin{figure}[!t]
    \centering
    \begin{adjustbox}{width=0.9\linewidth}    \includegraphics{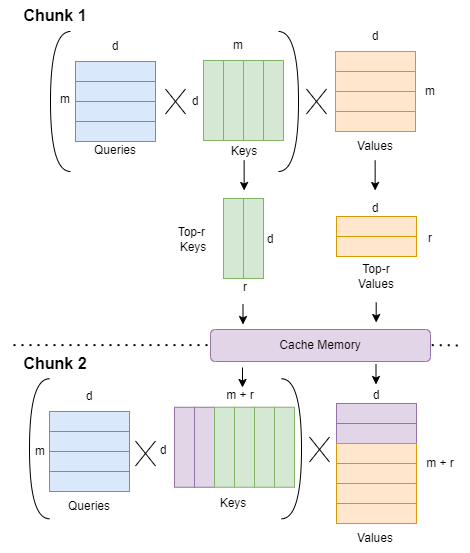}
   \end{adjustbox}
    \caption{Our attention computation process. In the top portion, the initial chunk \rev{of length $m$} is processed to identify the top \rev{$r$} keys,values pairs 
    through 
    the dot product of queries and keys. The top \rev{$r$} elements are then stored in cache memory. As the second chunk undergoes processing (bottom), new keys and values are generated 
    \rev{and both the current chunk of length $m$ and the top \rev{$r$} elements of the previous iteration are considered for the subsequent top \rev{$r$} selection.}
    } \label{fig:low-level}
\end{figure}

\vspace{0.5ex}
\noindent\textbf{Prompt-\rev{Guided} Layer-wise top \rev{r} position selection.} Our method for selecting the top \rev{$r$ (relevant)} positions is rooted in the analysis of the attention scores across its layers. We take into account the unique role of each layer for the representation of the input, i.e., early layers might focus on syntactic features, while deeper layers might capture more abstract, semantic relationships~\citep{clark-etal-2019-bert}. As a consequence, for each layer of the transformer, we calculate attention scores (the scaled dot-product attention between Q and K) and determine the context per-token relevance of the chunk with respect to tokens in the user prompt. By acknowledging that relevance varies by layer, we ensure a more holistic compression of the document. 
For example, tokens that are relevant in early layers might be not relevant in deeper layers.
This allows our method to preserve a wide spectrum of information 
without redundancy.

\begin{figure*}[ht]
\hspace{-4ex}
   \begin{adjustbox}{width=1.1\linewidth} 
\includegraphics{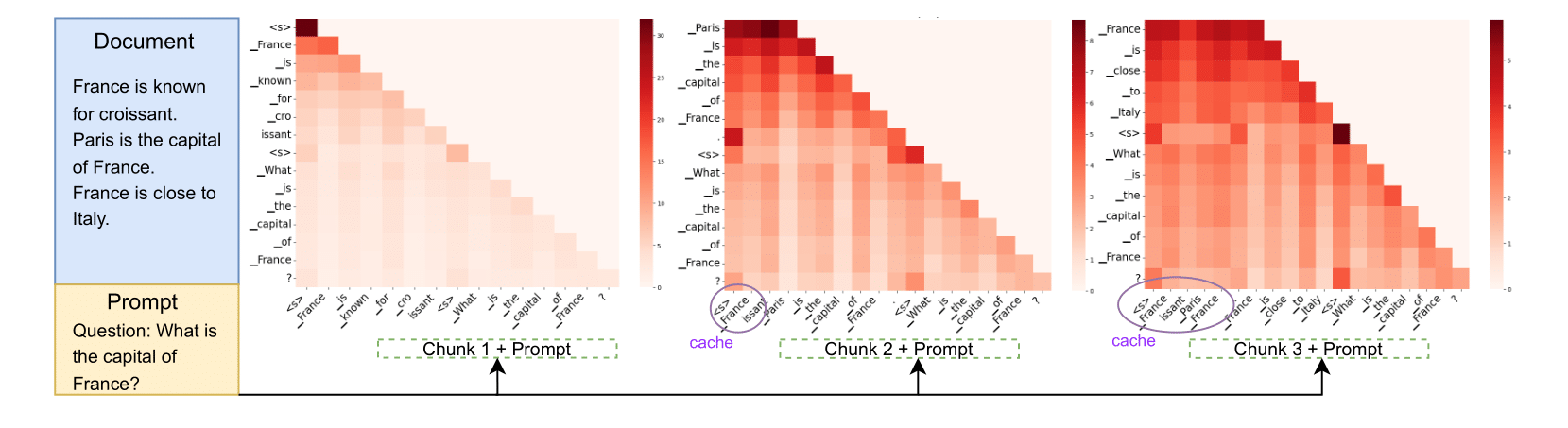}
   \end{adjustbox}
\caption{Attention distribution in the final layer of Llama 2, {\sc Finch} does the same analysis across all layers. The blue and yellow rectangles represent the document chunk and the user prompt, respectively.  
Initially, attention scores are evaluated between chunk 1 and the prompt, the most relevant tokens (circled) get stored in the cache in the next iteration. In successive iterations, the attention of the cached tokens together with the new chunk is measured w.r.t. the prompt. The final step involves only the cache and the prompt, leading the model to generate the response ``Paris'' based on the cached information.}
\label{fig:layer_visualization}
\end{figure*}

Our method also takes into consideration the inherent positional bias present in the attention mechanism. In particular, causal language models operate in the principle that each token in a sequence can only be influenced by preceding tokens, not by those that come after it. This is visually represented by a triangular matrix in attention mechanism, where the ability of tokens to "attend" to each other is constrained by their position in the sequence. As a result, early tokens in a sentence have a broader scope of attention compared to later tokens. For example, for the first token, its attention score is maximal since it only considers itself, leading to a score of 1. To address the issue that later tokens in the sequence, which could be equally or more relevant to the question, are not overlooked due to systemic bias, we incorporate a normalization step that adjusts the raw attention scores to mitigate positional bias, ensuring that each token's relevance is equally evaluated.

\rev{Consider \(\mathbf{A}^{(l)} \in \mathbb{R}^{H \times M \times N}\) as the attention scores matrix at layer \(l\), with \(H\) attention heads. Here, \(M\) and \(N\) are defined as: $$M = m + n^{\text{que}}, N = m + n^{\text{que}} + c$$
}
where $m$ is the chunk length and $c$ is the current KV cache length.
The compression process involves several steps as visualized in Figure \ref{fig:low-level}.

\begin{itemize}[leftmargin=*]
    \item \textbf{Sum over heads}: Every Head in a transformer attention layer captures various aspects of the data. 
    We sum the attention scores over the heads to aggregate their contributions, \rev{The elements \(A^{(l)\text{sum}}_{ij}\) of \(\mathbf{A}^{(l)\text{sum}}\) are defined as:} \rev{\begin{equation*}
\begin{split}
A^{(l)sum}_{ij} =& \sum_{h=1}^{H} A^{(l)sum}_{hij}\\
\forall \, i \in \{1, \ldots, M\},& j \in \{1, \ldots, N\}
\end{split}
\end{equation*}}
    \item \textbf{Extract prompt-\rev{guided} submatrix}: A submatrix is extracted to focus on the attention scores between prompt tokens and the current document chunk, this includes considering the tokens accumulated in the KV cache, which grows with each iteration:
    \rev{\begin{equation*}
\begin{split}
A^{(l)\text{cont}}_{i,j} =& A^{(l)\text{sum}}_{m+i,j} \\
\forall \, i \in \{1, \ldots, n^{\text{que}}\},& j \in \{1, \ldots, m + c \} \\
\end{split}
\end{equation*}}
Figure \ref{fig:layer_visualization} shows how attention scores for the last layer of Llama 2 evolve in the sequential operations.

    \item \textbf{Normalization}: Attention scores are normalized to mitigate positional bias, adjusting for non-zero attention scores:
    $$A^{(l)\text{norm}} = A^{(l)\text{cont}} \cdot \left(\frac{\text{count}(A^{(l)\text{cont}} \neq 0)}{m + c}\right)$$
    \item \textbf{Selection of top \rev{$r$} position}:
    The final step is to select the top \rev{$r$} indices based on the aggregated attention scores over the prompt tokens.
    \rev{
    \begin{equation*}
    \begin{split}
    A^{(l)\text{agg}}_{i} &= \sum_{p=1}^{n^{\text{que}}} A^{(l)\text{norm}}_{p,i} 
    \quad 
    \forall \, i \in \{1, \ldots, m+c\}
    \end{split}
\end{equation*}}
    \rev{$$\mathbf{t} = {\text{top-{r}}}(\mathbf{A^{(l)\text{agg}}}, r)$$}
    here, \rev{$\mathbf{t}$} is a vector containing indices of the top \rev{$r$} positions with the highest attention scores. 
    \rev{The parameter $r$ dynamically updates at each iteration based on the chunk size $m$, cache length $c$, and compression rate $\sigma$. Specifically, the update rule is given by:
    $$r_{it+1} = \frac{m_{it+1}}{\sigma} + c_{it}$$ where $it$ denotes the iteration. At the final iteration, $r$ corresponds to the target token size $k$.}

\end{itemize}

\vspace{0.5ex}
\noindent\textbf{Managing the Cache}:  
The key, value pairs for the selected top \rev{$r$} positions are preserved within the KV cache due to their significant relevance to the user prompt.
This process involves an adjustment to their positional embeddings. To accurately reflect the tokens' relative positions, we draw inspiration from the mechanisms used in Attention sinks~\citep{xiao2024efficient}. 
For example, given a cache sequence [0, 1, 2, 3, 4, 5] and a relevance ranking [3, 5, 0], we prioritize '3' by moving it three positions to the left, '5' by moving it four positions to the left, and '0' by shifting it two positions to the right, while the others are discarded. For Rotary Position Embeddings
~\citep{su2024roformer}, as in Llama 2, this repositioning involves calculating the cosine and sine required for rotating to earlier or later positions in the sequence.

\vspace{0.5ex}
\noindent\textbf{Compression Output}: The final cache, composed of $\mathbf{\Tilde{K}}$ and $\mathbf{\Tilde{V}}$, represents the compressed document, which encapsulates its essence in a condensed form and is used in the Generation stage.

\begin{table*}[ht]
\centering
\small
\begin{tabular*}{\textwidth}{@{\extracolsep{\fill}}llll}
\toprule
Method & Complexity per Layer & Sequential Ops & Cache Growth/Op. \\
\midrule
Vanilla & $O(n^2d)$ & $O(1)$ & $\Delta c = n$ \\
\textsc{Finch}   & $O(mcd + m^2d)$ & $O\left(\frac{n}{m}\right)$ & $\Delta c = \frac{m}{\sigma}$ \\
\bottomrule
\end{tabular*}
\caption{Complexity comparison between the Vanilla transformer and \textsc{Finch} in the Prefill stage. }
\label{tab:combined_attention_prefill}
\end{table*}

\begin{table*}[ht]
\centering
\small
\begin{tabular*}{\textwidth}{@{\extracolsep{\fill}}lllll}
\toprule
Method & Compl. per Layer & Seq. Ops & Initial Cache Size & Cache Growth/Op. \\
\midrule
Vanilla & $O(cd)$ & $O(a)$ & $c = n$ & $\Delta c = 1$ \\
\textsc{Finch}   & $O(cd)$ & $O(a)$ & $c = \frac{n}{\sigma}$ & $\Delta c = 1$ \\
\bottomrule
\end{tabular*}
\caption{Complexity comparison between the Vanilla transformer and \textsc{Finch} in the Generation stage.}
\label{tab:combined_attention_gen}
\end{table*}

\subsection{Complexity Analysis}
\label{subsec:complexity}
    To illustrate the computational benefit of our approach, we report a comparative analysis of complexity metrics between the attention-based Vanilla transformer and \textsc{Finch}. We consider Complexity per Layer according to $n$ (total number of tokens), $m$ (chunk size), $d$ (model's embedding dimension), $a$ (output sequence length), Sequential Operations as the number of times the model is invoked sequentially, Cache Growth per Operation as the increment in cache size $c$ with each sequential operation, and Initial Cache Size at the beginning of the Generation stage (0 at the beginning of the Prefill stage).
Table \ref{tab:combined_attention_prefill} shows complexities for the Prefill stage. 
For large $n$, the Vanilla method has a higher computational complexity due to quadratic relations, while \textsc{Finch} introduces sequential operations that scale according to $m$, hence demonstrating enhanced efficiency and potential for scalability in processing large sequences \rev{($m \ll n$)}.
Table \ref{tab:combined_attention_gen} shows complexities in the Generation stage, comparing the resource usage when synthesizing the final output. Also in this stage, the benefit for \textsc{Finch} come from the reduced size of the initial cache according to the compression ratio \rev{$\sigma$}.

\subsection{Encoder-decoder}
Our presentation of the methods is focused on a decoder-only architecture, as it is increasingly prevalent in NLP applications. While our methodology is experimented with decoder-only models, 
it is equally viable for encoder-decoder models that employ a KV cache mechanism. In such scenarios, during the Prefill stage, 
we can prefill the KV cache enabling the concise representation of context within the decoder. Subsequently, in the Generation stage we can feed the question or instructions to the encoder. The decoder then utilizes cross-attention mechanisms to access this information, along with the compressed context stored in the KV cache to generate the answer. 

\section{Experimental Setup}
We evaluate \textsc{Finch} using a variety of datasets and NLP tasks, with a focus on its application to the Llama 2 7B-chat~\citep{touvron2023llama2} \rev{and the Mistral 7B-Instruct-v0.2~\citep{jiang2023mistral} models}. 
Experiments are conducted 
with 
4-bit NormalFloat 
Quantization and Double Quantization
~\citep{dettmers2024qlora}. \rev{Unless otherwise noted, the experiments are conducted in a zero-shot setting.}\footnote{
\rev{\textsc{Finch}'s code and datasets are available at \url{https://anonymous.4open.science/r/context-compression-EAF6/README.md}}}.
Experiments are structured around three public datasets \rev{and four baseline methods}.\footnote{\rev{Details on the inference hyperparameters and on the chunk size $m$ per every dataset are reported in Tables \ref{tab:inference-hyperparameters} and \ref{tab:chunk-sizes}, respectively, in the Appendix.}
}

\vspace{0.5ex}
\noindent \textbf{SQuADv2}: For an assessment of \textsc{Finch}'s ability to preserve quality when compressing according to Equation~\ref{eq:similar_predictions}, we use short texts that let us run the entire document as input. We use SQuAD v2~\citep{rajpurkar-etal-2018-know}, a benchmark which includes both questions that can and cannot be answered with the given documents. We measure how our model maintains or improves its accuracy, despite having reduced context, against two baselines. First, we report for \textit{Vanilla}, the standard model configuration which has access to the full context. \rev{Second, a \textit{Truncate} strategy that reduces the input to the same size used by \textsc{Finch}. Given a budget, we truncate the input after a number of tokens equal to half the reduced context both from the start and from the end, i.e., we take the beginning and the end of the document}. 

\vspace{0.5ex}
\noindent \textbf{LongBench}: To assess the robustness of our method with long documents and a variety of tasks, we also evaluate on the LongBench benchmark~\citep{bai2023longbench}. This is a suite of tasks that involve extended contexts, including single-document question answering (QA), multi-document QA, document summarization, few-shot learning, code completion, and  a synthetic task. The tasks span 16 datasets and presents a challenge due to the length of the \rev{input} texts; \rev{for the size of the output, we use the original values in the dataset (see Table~\ref{tab:sequence-lengths} in the Appendix for details)}. For this dataset, our model is also compared against \rev{a third baseline}, LongLLMLingua~\citep{jiang2024longllmlingua}, a state-of-the-art method for compression \rev{of long input} texts. 
For LongLLMLingua, we use phi-2~\citep{li2023textbooks} as the compressor and Llama 2 7B-chat, quantized at 4 bits with double quantization, as the generator. Unlike LongLLMLingua, our method does not use an external model for compression.
\rev{For question answering tasks, a natural baseline is a Retrieval Augmented Generation (\textsc{RAG})  solution~\citep{rag20}.
In our implementation of RAG, we segment the long text into chunks of 256 tokens each. To identify the most relevant chunks, we calculate the cosine similarity between the embeddings of these chunks and the embedding of the prompt. We use the \textit{all-mpnet-base-v2} model from Sentence Transformers~\citep{reimers2019sentence} for generating these embeddings.}

\vspace{0.5ex}
\noindent \textbf{Lost in the Middle}: A critical challenge for LLMs is the "lost in the middle" issue~\citep{tacl_a_00638}, where models exhibit degraded performance if relevant information is situated in the middle of long contexts. We evaluate the robustness of our compression technique also in their dataset.

\section{Results and Discussion}
\label{sec:results}
We discuss five questions over our results.

\vspace{0.5ex}
\noindent\textbf{1. Does {\sc Finch}'s \rev{compression} preserve the relevant information?}
Our evaluation on SQuAD v2 measures how {\sc Finch} retains pertinent information in a compressed format. We compare the Vanilla approach (Llama 2 provided with full documents), 
{\sc Finch} constrained to target tokens size $k$, 
and the truncation strategy. 
\rev{We choose five values of} target tokens sizes, corresponding to different average compression ratios; we 
obtain the latter by dividing the average number of tokens in the SQuAD tests (document and prompt) by the average number of tokens that {\sc Finch} uses according to the given target tokens size. \rev{Specifically, 384 target tokens corresponds to an average $\sigma$ of 1.1x, 256 tokens to 1.53x, 192 tokens to 2.35x, 160 to 3.03x and 144 tokens to 3.76x.} 

\begin{figure}[t]
\centering
\includegraphics[width=\linewidth]{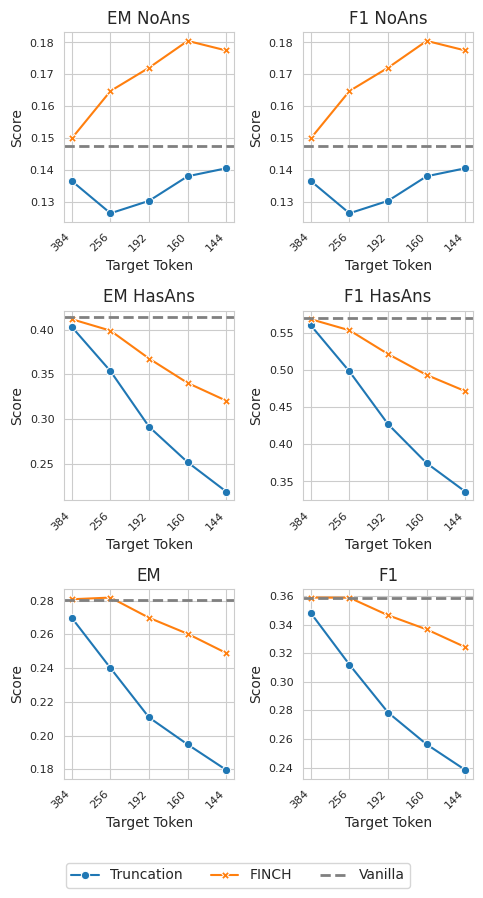}
\caption{Performance results for SQuAD v2 for the Llama 2 Vanilla model, {\sc Finch}, and the truncation baseline. We report Exact Match (EM) and F1 scores for tests without answers (top), tests with answers (middle) and average across all tests.}
\label{fig:squad}
\end{figure}

The results in Figure \ref{fig:squad} show that {\sc Finch} not only consistently outperforms the truncation strategy across all token lengths but also, in certain cases, exceeds the quality performance of the Vanilla approach. This is evident in the F1 NoAns and Exact Match (EM) NoAns scores, where {\sc Finch}'s ability to prevent responses based on irrelevant or non-existent evidence 
suggests that it eliminates extraneous content that could potentially mislead the model.

The overall EM and F1 scores indicate that {\sc Finch} 
maintains the integrity of the context as it is compressed. Even as the target tokens size $k$ decreases, {\sc Finch} holds onto essential information, enabling the model to generate accurate responses with significantly less input data. In this dataset, the loss of quality compared to the full context becomes more significant starting with an average compression of 3.7x.

\begin{table}[tbp]
\centering
\resizebox{\columnwidth}{!}{
\begin{tabular}{lccccc}
\hline
\textbf{Model} & \textbf{Idx 0} & \textbf{Idx 4} & \textbf{Idx 9} & \textbf{Idx 14} & \textbf{Idx 19} \\
\hline
Vanilla & 24.7\% & 25.2\% & 28.2\% & 29.7\% & 40.0\% \\
{\sc Finch} & 38.0\% & 36.4\% & 38.2\% & 41.1\% & 46.2\% \\
\hline
\end{tabular}
}
\caption{\rev{``Lost in the middle'' comparison of {\sc Finch} and Vanilla (Llama 2). Accuracy of returning the correct answer when the position of the document containing it varies across the model's input} ($n=4096$, $m=256$). {\sc Finch}'s 
$\sigma = 4$.}
\label{tab:lost_in_the_middle}
\end{table}

To further illustrate the impact of our compression, we run the ``lost in the middle'' experiment, where the position of the information to answer the user question changes within the input document. It has been shown that this position has a significant impact on the model's accuracy~\citep{tacl_a_00638}. We compare again our solution against the original Vanilla model \rev{on the dataset from the paper reporting this problem}.
Results in Table \ref{tab:lost_in_the_middle} show that {\sc Finch} significantly outperforms the baseline across the different positions, with up to 13.3 absolute points gain when the correct answer is in the first document \rev{(Idx 0)} \rev{and the compression ratio is 4x}. The results also show that our method mitigates the original ``lost in the middle'' issue with 9.8 absolute points difference between the best and worst accuracy for {\sc Finch}, rather than 15.3 points for Vanilla.

\begin{figure}[t]
\centering
\includegraphics[width=\linewidth]{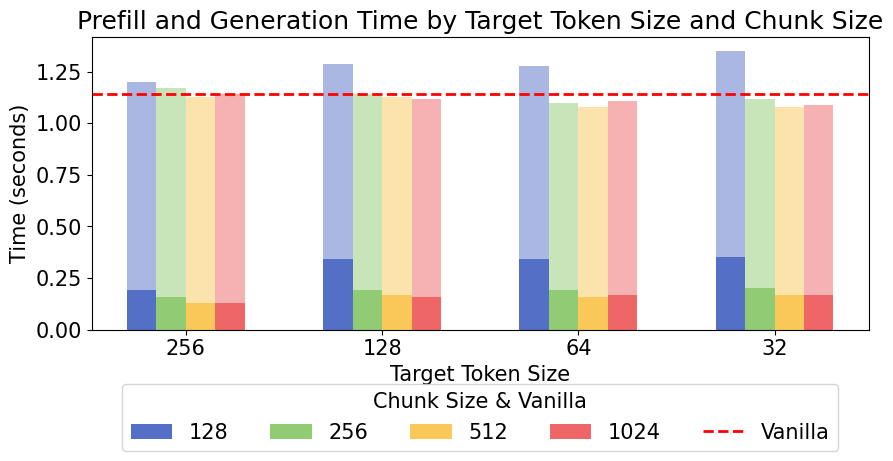}
\caption{Impact of chunk and target tokens size $k$ on decoding time for SQuAD v2; Finch's prefill (dark color) and generation (light color) times vs Llama 2 Vanilla (dotted red line: prefill+generation).}
\label{fig:squad_v2_time}
\end{figure}

\setlength{\tabcolsep}{2.5pt}
\begin{table*}
\centering
\begin{adjustbox}{width=\textwidth,center}
\begin{tabular}{@{}lllcccccccccccc@{}}
\toprule
& & & \multicolumn{4}{c}{512 target tokens} & \multicolumn{4}{c}{1000 target tokens} & \multicolumn{4}{c}{2000 target tokens} \\  \cmidrule(l{4pt}r{4pt}){4-7} \cmidrule(l{4pt}r{4pt}){8-11} \cmidrule(l{4pt}r{4pt}){12-15}
Task (metric) & Dataset & Vanilla & Truncate & \textsc{Finch} & \textsc{Lingua} 
& $\text{avg}(\sigma)$ &  Truncate & \textsc{Finch} & \textsc{Lingua} 
& $\text{avg}(\sigma)$ &  Truncate & 
 \textsc{Finch} & \textsc{Lingua} 
 & $\text{avg}(\sigma)$  \\
\midrule
\multirow{3}{*}{
  \begin{tabular}{@{}c@{}}Single-Doc QA \\ (F1 score $\uparrow$)\end{tabular} 
} & NarrativeQA 
& 21.64 & 9.84 & 17.85 & 9.13 
& 77.50x & 11.28 & 20.38 & 9.13 
& 37.16x & 14.72 &  17.60 & 9.24 
& 18.1x\\
 & Qasper
 & 24.93 & 9.23 & 19.59 & 9.71 
 & 13.68x & 12.52 & 22.18 & 12.36 
 & 6.51x & 16.50 &  23.19 & 15.62 
 & 3.40x \\
 & MultiFieldQA 
 & 45.13 & 29.56 & 37.47 & 23.31 
 & 16.7x & 36.8 &  42.11 & 24.60 
 & 8.52x& 41.44 &  44.13 & 29.70 
 & 4.42x \\
 \cmidrule{2-15}
& Overall & 30.57 & \textit{16.21} & \textbf{24.97} & 14.05 
& & \textit{20.20} &  \textbf{28.22} & 15.36 
& & \textit{24.22} & \textbf{28.30} & 18.18 
&\\
\midrule
\multirow{3}{*}{
 \begin{tabular}{@{}c@{}}Multi-Doc QA \\ (F1 score $\uparrow$)\end{tabular} 
 } & HotpotQA 
 & 17.15 & 19.20 & 29.89 & 18.28 
 & 34.38x & 22.62 & 33.41 & 18.91 
 & 16.81x & 26.43 &  33.21 & 25.01 
 & 8.42x  \\
 & 
 MultihopQA 
 & 21.65 & 13.62 & 16.17 & 12.51 
 & 19.63x & 14.79 & 18.42 & 13.74 
 & 9.85x & 16.26 & 25.28 & 14.15 
 & 5.20x \\
 & MuSiQue 
 & 19.25 & 7.58 & 12.43 & 6.09 
 & 39.96x & 9.23 & 15.7 & 6.47 
 & 19.40x & 11.94 & 17.86 & 8.23 
 & 9.64x \\
 \cmidrule{2-15}
& Overall & 19.35 & \textit{13.47} & \textbf{19.49} & 12.29 
&& \textit{15.55} & \textbf{22.51} & 13.09 
&& \textit{18.21} & \textbf{25.45} & 15.80 
&\\
\midrule
\multirow{3}{*}{
 \begin{tabular}{@{}c@{}}Summarization\\ (Rouge-L $\uparrow$)\end{tabular} } & GovReport 
 & 24.24 & 18.70 & 19.05 & 18.16 
 & 25.1x & 20.07 & 20.12 & 18.46 
 & 12.64x & 21.36 & 21.05 & 19.03 
 & 6.50x \\
 & QMSum 
 & 20.52 & 17.95 & 19.86 & 18.20 
 & 33.84x & 18.86 & 20.04 & 18.03 
 & 16.72x & 18.80 & 20.08 & 18.43 
 & 8.52x \\
 & MultiNews & 18.58 & 16.85 & 16.95 & 16.39 
 & 7.32x & 17.94 & 17.79 & 16.89 
 & 3.89x & 18.47 & 18.31 & 18.40 
 & 2.16x \\
  \cmidrule{2-15}
& Overall & 21.11 & \textit{17.83} & \textbf{18.62} & 17.58 
&& \textit{18.96} & \textbf{19.31} & 17.79 
&& \textit{19.54} & \textbf{19.81} & 18.62 
& \\
\midrule
\multirow{2}{*}{
 \begin{tabular}{@{}c@{}}Few-shot Learn\\ (Accuracy $\uparrow$)\end{tabular} } 
  & TREC & 29.79 & \textbf{40.39} & \textit{36.75} & 17.17 
  & 16.47x & \textit{43.14} & \textbf{43.68} & 10.08 
  & 8.37x & \textit{44.43} & \textbf{47.41} & 16.62 
  & 4.47x \\
  &  &  &  &  &  & & & & & \\
\midrule
\multirow{2}{*}{
 \begin{tabular}{@{}c@{}}Synthetic Task\\ (Accuracy $\uparrow$)\end{tabular} } 
   & PassageCount & 0.96 & 0.25 & \textit{1.35} &\textbf{3.00} 
   & 41.61x & 0.96 & \textbf{2.41} & \textit{2.00} 
   & 19.08x & \textit{2.25} & \textbf{2.81} & 2.21 
   & 9.33x \\
 &  &  &  &  &  &  &  & & & \\
\midrule
\multirow{2}{*}{
 \begin{tabular}{@{}c@{}}Code Complete \\ (Edit Sim $\uparrow$)\end{tabular} 
 } & LCC & 26.01 & 18.97 & 31.93 & 15.08 
 & 9.78x & 22.74 & 33.34 & 15.55 
 & 5.20x & 24.31 & 34.59 & 18.56 
 & 2.91x \\
 & RepoBench-p & 25.65 & 18.51 & 24.19 & 15.64 
 & 28.65x & 21.21 & 25.26 & 16.46 
 & 14.62x & 23.34 & 25.63 & 18.60 
 & 7.55x \\
  \cmidrule{2-15}
& Overall & 25.83 & \textit{18.74} & \textbf{28.06} & 15.36 
&& \textit{21.98} & \textbf{29.30} & 16.01 
&& \textit{23.83} & \textbf{30.11} & 18.58 
&\\
\bottomrule
\end{tabular}
\end{adjustbox}
\caption{\rev{Mistral results' comparison for the full context (Vanilla), truncation (Truncate), \textsc{Finch} and LongLLMLingua (\textsc{Lingua}) compression for different target tokens sizes (512/1000/2000) across datasets for six tasks. Best result per task and target tokens size in \textbf{bold}, second best in \textit{italic}.}}
\label{tab:mistral_revised_comparison}
\end{table*}

\setlength{\tabcolsep}{2.5pt}
\begin{table*}
\centering
\begin{adjustbox}{width=\textwidth,center}
\begin{tabular}{@{}lllcccccccccccc@{}}
\toprule
& & & \multicolumn{4}{c}{512 target tokens} & \multicolumn{4}{c}{1000 target tokens} & \multicolumn{4}{c}{2000 target tokens} \\  \cmidrule(l{4pt}r{4pt}){4-7} \cmidrule(l{4pt}r{4pt}){8-11} \cmidrule(l{4pt}r{4pt}){12-15}
Task (metric) & Dataset & Vanilla & Truncate & \textsc{Finch} & \textsc{Lingua} 
& $\text{avg}(\sigma)$ &  Truncate & \textsc{Finch} & \textsc{Lingua} 
& $\text{avg}(\sigma)$ &  Truncate & 
 \textsc{Finch} & \textsc{Lingua} 
 & $\text{avg}(\sigma)$  \\
\midrule
\multirow{3}{*}{
  \begin{tabular}{@{}c@{}}Single-Doc QA \\ (F1 score $\uparrow$)\end{tabular} 
} & Narrative
& 16.69 & 11.14 & 19.10 & 10.56 
& 93.17x & 14.15 &   18.15 & 10.51 
& 40.92x & 15.45 &  19.45 & 11.68 
& 19.37x\\
 & Qasper
 & 12.53 & 11.81 & 19.39 & 12.10 
 & 15.62x & 12.27 & 20.25 & 11.82 
 & 7.00x & 12.78 &  22.95 & 12.70 
 & 3.46x \\
 & MultiField
 & 34.50 & 30.26 & 33.47 & 21.87 
 & 17.86x & 32.67 &  33.88 & 23.18 
 & 8.85x& 38.43 &  34.67 & 27.35 
 & 4.50x \\
 \cmidrule{2-15}
& Overall & 21.24 & \textit{17.74} & \textbf{23.99} & 14.84 
& & \textit{19.70} &  \textbf{24.09} & 15.17 
& & \textit{22.22} & \textbf{25.69} & 17.24 
&\\
\midrule
\multirow{3}{*}{
 \begin{tabular}{@{}c@{}}Multi-Doc QA \\ (F1 score $\uparrow$)\end{tabular} 
 } & Hotpot
 & 30.46 & 25.31 & 36.75 & 26.13 
 & 38.64x & 29.47 & 36.48 & 27.29 
 & 17.90x & 30.07 &  34.29 & 28.32 
 & 8.71x  \\
 & 
 Multihop
 & 26.47 & 22.04 & 28.81 & 25.34 
 & 21.07x & 22.90 & 27.96 & 24.64 
 & 10.24x & 26.78 & 30.22 & 25.72 
 & 5.13x \\
 & MuSiQue
 & 10.54 & 9.41 & 14.12 & 9.43 
 & 45.97x & 9.41 & 13.93 & 9.61 
 & 20.66x & 8.25 & 12.58 & 10.21 & 10.03x 
 \\
 \cmidrule{2-15}
& Overall & 22.49 & 18.92 & \textbf{26.56} & \textit{20.30} 
&& \textit{20.59} & \textbf{26.12} & 20.51 
&& \textit{21.70} & \textbf{25.10} & 21.42 
&\\
\midrule
\multirow{3}{*}{
 \begin{tabular}{@{}c@{}}Summarization\\ (Rouge-L $\uparrow$)\end{tabular} } & GovReport
 & 18.02 & 17.79 & 18.20 & 17.27 
 & 28.30x & 18.61 & 18.41& 17.32 
 & 13.73x & 19.19 & 18.79 & 17.86 
 & 6.84x \\
 & QMSum
 & 19.29 & 18.41 & 19.80 & 19.01 
 & 37.02x & 18.47 & 19.63 & 18.86 
 & 17.38x & 19.56 & 19.99 & 19.37 
 & 8.74x \\
 & MultiNews
 & 16.70 & 16.89 & 16.57 & 15.97 
 & 7.82x & 17.29 & 17.22 & 16.61 
 & 4.11x & 17.62 & 17.52 & 17.57 
 & 2.23x \\
  \cmidrule{2-15}
& Overall & 18.00 & \textit{17.70} & \textbf{18.19} & 17.42 
&& \textit{18.12} & \textbf{18.42} & 17.60 
&& \textbf{18.80} & \textit{18.77} & 18.26 
& \\
\midrule
\multirow{2}{*}{
 \begin{tabular}{@{}c@{}}Few-shot Learn\\ (Accuracy $\uparrow$)\end{tabular} } 
  & TREC & 15.00 & \textbf{24.25} & \textit{23.75} & 6.50 
  & 17.75x & \textit{25.00} & \textbf{26.00} & 6.50 
  & 8.78x & \textbf{32.50} & \textit{29.00} & 8.00 
  & 4.57x \\
  &  &  &  &  &  &  &  &  & & \\
\midrule
\multirow{2}{*}{
 \begin{tabular}{@{}c@{}}Synthetic Task\\ (Accuracy $\uparrow$)\end{tabular} } 
   & P. Count 
   & 4.25 & \textbf{5.17} & 2.45 & \textit{4.50} 
   & 43.58x & \textbf{3.17} & 2.32 & \textit{3.00} 
   & 19.65x & \textbf{2.60} & 1.67 & \textit{2.00} 
   & 9.52x \\
 &  &  &  &  &  &  &  &  & & \\
\midrule
\multirow{2}{*}{
 \begin{tabular}{@{}c@{}}Code Complete \\ (Edit Sim $\uparrow$)\end{tabular} 
 } & LCC & 21.16 & 25.52 & 26.02 & 25.02 
 & 10.21x & 25.06 & 25.79 & 22.14 
 & 5.32x & 24.64 & 
24.64 & 20.45 
& 2.98x \\
 & R. Bench 
 & 23.00 & 24.23 & 25.88 & 26.73 
 & 29.84x & 23.33 & 24.67 & 24.11 
 & 14.97x & 23.34 & 23.46 & 21.14 
 & 7.65x \\
  \cmidrule{2-15}
& Overall & 23.28 & 24.88 & \textbf{25.95} & \textit{25.88} 
&& \textit{24.20} & \textbf{25.23} & 23.13 
&& \textit{24.00} & \textbf{24.05} & 20.80 
&\\
\bottomrule
\end{tabular}
\end{adjustbox}
\caption{\rev{Llama 2 results' comparison for the full context (Vanilla), truncation (Truncate), \textsc{Finch} and LongLLMLingua (\textsc{Lingua}) compression for different target tokens sizes (512/1000/2000) across datasets for six tasks. Best result per task and target tokens size in \textbf{bold}, second best in \textit{italic}.}}
\label{tab:llama_revised_comparison}
\end{table*}

\vspace{0.5ex}
\noindent\textbf{2. How fast is {\sc Finch} compared to Vanilla self attention?}
Analysis of {\sc Finch}'s efficiency, detailed in Figure \ref{fig:squad_v2_time}, highlights a reduction on the overall time w.r.t. the Vanilla when the chunk size is greater than 128 on Llama 2. This observation aligns with the complexity study in Section~\ref{subsec:complexity}. Although {\sc Finch} introduces additional sequential operations in the Prefill stage, these are offset by the reduced complexity per layer, which is contingent on the chunk size $m$ rather than the full context size $n$. This approach allows {\sc Finch} to handle each chunk with a complexity of $O(mcd + m^2d)$ as opposed to the Vanilla complexity per layer $O(n^2d)$. With larger chunk sizes, {\sc Finch} demonstrates improved speed over Vanilla self-attention. In the generation phase, the distinction in performance becomes more pronounced, as in Table \ref{tab:combined_attention_gen}. {\sc Finch} benefits from a smaller initial cache size, which is a function of the compression ratio $\sigma$. Such a configuration is advantageous in real-world applications where the response time is key and the volume of text to be processed is substantial.

\vspace{0.5ex}
\noindent\textbf{3. How \rev{does {\sc Finch} perform} on documents larger than the model context?}
To study how our method handles long \rev{input} documents, we focus on the LongBench benchmark. \rev{As for the SQuADv2 experiment, we set the target tokens sizes and we feed the input document in chunks, while reserving space for the prompt and the output generation.}
We compare {\sc Finch} also against the state-of-the-art compression model LongLLMLingua.\footnote{Results for LongLLMLingua are lower than those reported in their paper, where they use larger models such as ChatGPT~\citep{jiang2024longllmlingua}.} As shown in Table~\ref{tab:mistral_revised_comparison} \rev{and Table~\ref{tab:llama_revised_comparison}}, {\sc Finch} outperforms LongLLMLingua across five of the six tasks \rev{on Mistral and four out of six on Llama 2}. The \rev{benefit} of our solution is clear with different datasets and compression ratios, with \rev{a boost up to 8.8 absolute points of accuracy for question answering w.r.t. the best baseline (Truncate) on Mistral. Experiments on Llama 2 reports similar patterns, with a an improvement up to 6.3 
points over the best QA baseline.} 

\rev{{\sc Finch} outperforms also the Vanilla baseline using the full document as input in the model context in 12 of the 18 experiments (overall results across 6 tasks and 3 target tokens sizes) on Mistral and in 15 over 18 on Llama 2. This is remarkable when considering that the compression ratio varies between 2.23x and 93.17x.}

\rev{The baselines beat our method in 4 out of 6 experiments in the Synthetic task, where all methods report very low results.} We explain this by the limits of the LLM with 7B parameters, since the tasks demands deep contextual understanding. 

{\sc Finch} shows better performance according to increasing target tokens sizes (512, 1000, 2000). In the
\rev{question answering tasks, {\sc Finch} with a compression at 512 target tokens beats Truncate and LongLLMLingua with 1000 and 2000 target tokens, both with Llama 2 and Mistral}.

\begin{figure}[t]
\centering
\includegraphics[width=\linewidth]{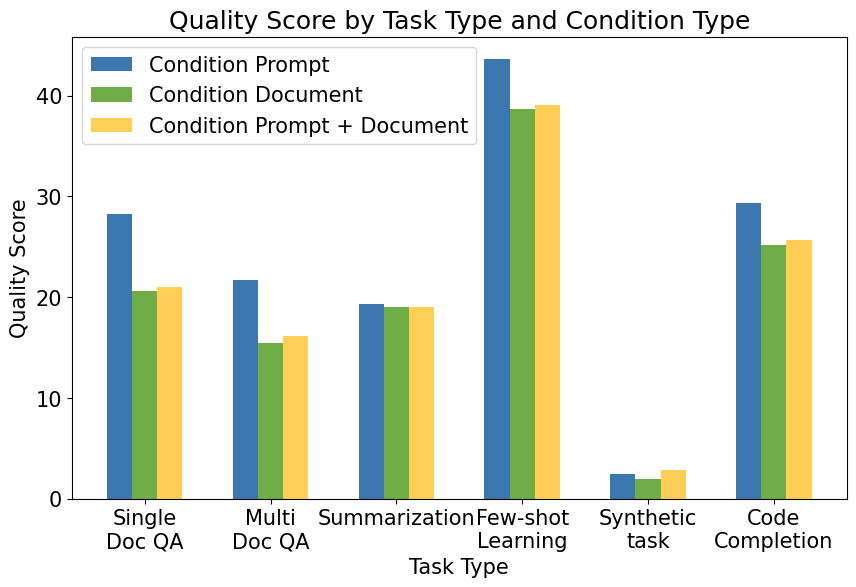}
\caption{\rev{Impact of three types of condition in {{\sc Finch}} in all LongBench tasks on Mistral.}}
\label{fig:ablation_condition_new}
\end{figure}

\begin{figure*}[ht]
\centering
\includegraphics[width=.78\linewidth]{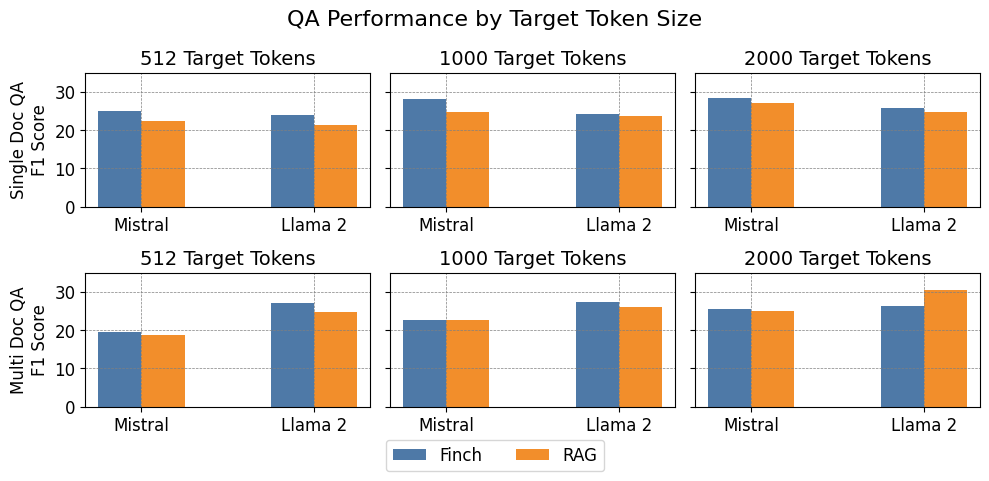}
\caption{\rev{Comparison of \textsc{Finch} and \textsc{RAG} in Mistral and Llama 2 for the QA tasks of LongBench.}}
\label{fig:finch_vs_rag}
\end{figure*}

\rev{We use the LongBench datasets also to validate our idea that conditioning the compression guided by the prompt is more effective than analyzing the self attention scores on the entire input (prompt and document) or on the document only. Results in Figure~\ref{fig:ablation_condition_new} show that over all the six tasks, the prompt guided solution leads to the best quality.}

\begin{figure*}[ht]
\centering
\includegraphics[width=\linewidth]{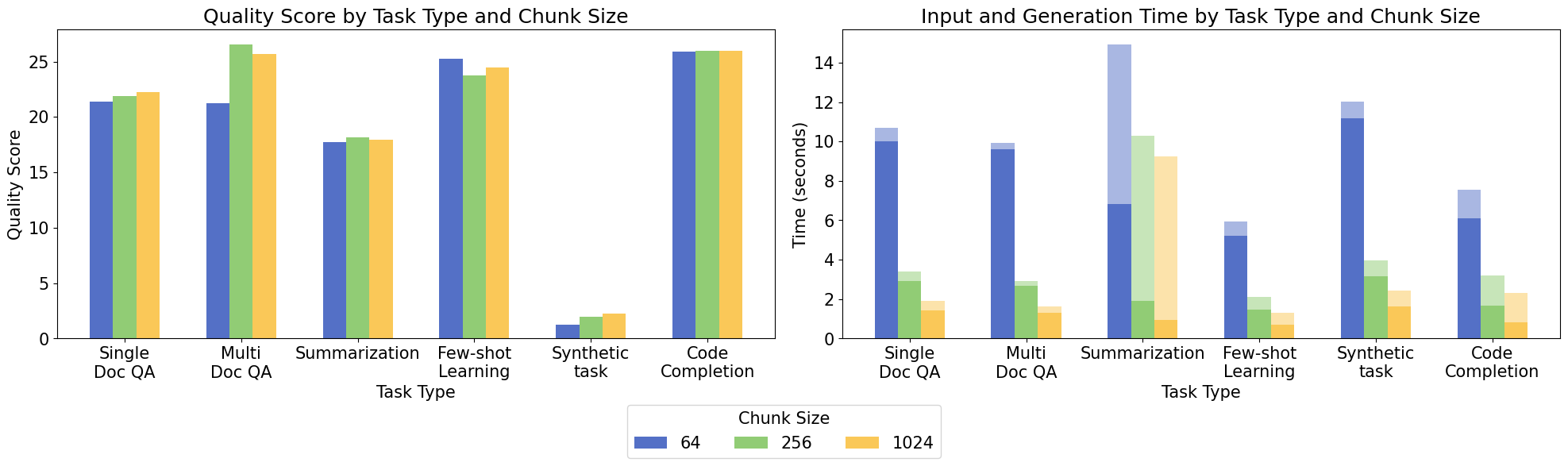}
\caption{Ablation study for the impact of three chunk sizes in \rev{{\sc Finch}} in all LongBench tasks \rev{on Llama 2} \rev{with a target tokens size of 512}. Left: quality score. Right: inference Prefill (dark color) and Generation (light color) execution times.}
\label{fig:ablation}
\end{figure*}

\rev{Finally, Figure~\ref{fig:finch_vs_rag} shows how {\sc Finch} outperforms the {\sc RAG} baseline both on Mistral and Llama 2 at different compression rates in 10 over 12 question answering experiments. Compressing with {\sc Finch}, using the LLM KV cache, offers superior reliability w.r.t. a RAG solution, which suffers from increased latency and fragility due to its dependency on external retrieval mechanisms.}

\vspace{0.5ex}
\noindent\textbf{4. What is the effect of the chunk size?}
Figure~\ref{fig:ablation} shows the impact of the chunk size $m$, i.e., the number of tokens into which the \rev{input} context is divided for sequential processing by the model. Results show nuanced effects on quality performance. Larger chunk sizes (1024) yield better performance in single-document question answering, while smaller sizes (256) are more effective in multi-document settings. This can be attributed to the compression algorithm of retrieving a fixed number of top \rev{r} tokens per iteration. In noisy multi-document contexts, a smaller chunk size enables better discrimination between relevant and irrelevant content, enhancing overall model performance.
Chunk size has also an impact on the execution times. As expected, larger chunks lead to faster end-to-end execution because of the smaller number of iterations. \rev{These positive results are especially important for use cases that require longer outputs generated by the LLMs. As the user requires a bigger output, the space available for input processing gets smaller, thus reducing the size of the chunks in the Prefill stage.}

\begin{table}[htbp]
\centering
\small
\begin{tabular}{lcc}
\hline
\textbf{Method} & \multicolumn{1}{p{2cm}}{\centering \textbf{Model (GB)}} & \multicolumn{1}{p{2.5cm}}{\centering \textbf{KV Cache (GB)}} \\ \hline
Vanilla        & 4.33 & 4.52 \\
{\sc Finch} ($\sigma=2$) & 4.33 & 2.38 \\
{\sc Finch} ($\sigma=4$) & 4.33 & 1.30 \\
{\sc Finch} ($\sigma=8$) & 4.33 & 0.60 \\ \hline
\end{tabular}%
\caption{Memory consumption of Vanilla and {\sc Finch} at the beginning of the Generation stage.}
\label{tab:memory_consumption}

\end{table}

\noindent\textbf{5. What is the benefit in terms of GPU memory?}
Table \ref{tab:memory_consumption} reports the memory consumed by {\sc Finch} (different compression rates) and the Vanilla model for the NarrativeQA (LongBench) dataset (truncated at $n=4096$). Results show that our approach delivers a significant reduction in the initial KV cache size at the beginning of the Generation stage. Unlike the Vanilla model, 
{\sc Finch} achieves substantial memory savings by reducing the required cache size in proportion to the compression ratio, confirming the 
 results in Table \ref{tab:combined_attention_gen}. This benefit enhances model scalability and makes {\sc Finch} a practical choice for deployment in resource-constrained environments.

\section{Conclusion and Future Work}
We have shown how attention can be used to identify and prioritize important information within the input data, effectively reducing the need for truncation.
\textsc{Finch} tackles the limitations of LLMs in processing large inputs, offering a balance between computational efficiency and maintaining high language model quality. Our solution leverages the pre-trained model weights of the self-attention mechanism to provide an economically feasible method for operating LLMs.

As future work, we envision a dynamic threshold mechanism to avoid that a fixed amount of KV states are selected in every chunk of the Prefill stage, exploiting the fact that some chunks are not relevant and can be compressed more. 
\rev{Another interesting research question is about the use of the proposed method to compress the generated output tokens. This extension would be especially valuable in settings where the LLM is requested to generate long outputs, such as chain-of-thought reasoning. Our approach could be used to identify the important tokens to preserve in the generation step - this is aligned with results showing that preserving a fraction of the original context is sufficient to obtain high quality generated outputs~\citep{xiao2024efficient,han2024lminfinite}.}

Finally, we are interested in studying how cache compression techniques can be extended to structured data, e.g., for replacing the current data retrieval and filtering solution in table question answering~\citep{Badaro0P23}.

\noindent\textbf{Acknowledgments.} We thank the action editor and the reviewers for their comments which improved the content of this work. We also thank Riccardo Taiello for the insightful discussion on complexity analysis.

\bibliography{tacl2021}
\bibliographystyle{acl_natbib}

\newpage
\appendix
\section*{Appendix}
\label{sec:appendix}

\begin{table}[h]
    \centering
        \small
    \begin{tabular}{ll}
        \toprule
        Hyperparameter & Value \\
        \midrule
        Number of Beams & 1 \\
        Do Sample & False \\
        Temperature & 1.0 \\
        Top-k & 50 \\
        Top-p & 1.0 \\
        Repetition Penalty & 1.0 \\
        \bottomrule
    \end{tabular}
    \caption{\rev{Inference hyperparameters used for Llama 2 and Mistral.}}
    \label{tab:inference-hyperparameters}
\end{table}

\begin{table}[h]
    \centering
    \small
    \begin{tabular}{ll}
        \toprule
        Symbol & Description \\
        \midrule
        \( m \)    & Chunk size   \\
        \( n \)    & Total sequence length (context and prompt) \\
        \( n^{\text{que}} \) & Sequence length of prompt           \\
        \( n^{\text{cont}} \) & Sequence length of context          \\
        \( a \)    & Output sequence length               \\
        \( d \)    & Embedding dimension                  \\
        \( \sigma \) & Compression factor                  \\
        \( k \)    & Target tokens                        \\
        \( c \)    & Current cache length                 \\
        \( r \)    & Current relevant tokens              \\
        \( n_{max} \)    & Maximum model's context window\\
        \( m_{max} \)    & Maximum model's chunk size\\
        \( \mathbf{x^{\text{cont}}} \)    &  Context sequence\\
        \( \mathbf{x^{\text{que}}} \)    & Prompt sequence\\
        \( \mathbf{x} \) & Total input sequence (context and prompt) \\
        \( \mathbf{y} \) & Output sequence  \\
        \( \mathbf{\tilde{y}} \) & Output sequence  using compressed context\\
        \( \mathbf{K, V} \) & Key, Value matrices  \\
        \(\mathbf{\Tilde{K}},\mathbf{\Tilde{V}}\) & Compressed Key, Value matrices \\
        \( \mathbf{K^{\text{cont}}, V^{\text{cont}}} \) & Context Key, Value matrices  \\
        \( \mathbf{A} \) & Attention matrix  \\

        \bottomrule
    \end{tabular}
    \caption{\rev{Summary of symbols used in this work.}}
    \label{table:notation}
\end{table}
\begin{table}[h]
    \centering
    \small
    \begin{tabular}{lcc}
        \toprule
        Dataset & Llama 2 & Mistral \\
        \midrule
        Lost-in-the-Middle & 256 & - \\
        SQuAD v2 & 512 & - \\
        Narrative & 256 & 2048 \\
        Qasper & 64 & 2048 \\
        MultiField & 1024 & 2048 \\
        Hotpot & 256 & 2048 \\
        MultiHop & 256 & 2048 \\
        MuSiQue & 256 & 512 \\
        GovReport & 256 & 2048 \\
        QMSum & 256 & 2048 \\
        MultiNews & 256 & 2048 \\
        TREC & 256 & 2048 \\
        P. Count & 2048 & 2048 \\
        LCC & 256 & 2048 \\
        R. Bench & 1024 & 2048 \\
        \bottomrule
    \end{tabular}
    \caption{\rev{Chunk size $m$ values used per dataset for Llama 2 and Mistral.}}
    \label{tab:chunk-sizes}
\end{table}

\begin{table}[h!]
    \centering
        \small
    \begin{tabular}{lcc}
        \toprule
        Dataset & Max Prompt  & Max Answer  \\
         &   Length &   Length \\
        \midrule
        Lost-in-the-Middle & 128 & 100 \\
        SQuAD v2 & 128 & 32 \\
        Narrative & 128 & 128 \\
        Qasper & 256 & 128 \\
        MultiField & 128 & 256 \\
        Hotpot & 128 & 32 \\
        MultiHop & 128 & 32 \\
        MuSiQue & 128 & 32 \\
        GovReport & 128 & 512 \\
        QMSum & 128 & 512 \\
        MultiNews & 128 & 512 \\
        TREC & 128 & 64 \\
        P. Count & 256 & 32 \\
        LCC & 128 & 64 \\
        R. Bench & 128 & 64 \\
        \bottomrule
    \end{tabular}
    \caption{\rev{Maximum prompt and answer sizes for each dataset. For the instruction prompt, we used those reported in \citep{bai2023longbench}}.}
    \label{tab:sequence-lengths}
\end{table}

\end{document}